# Radio frequency detection and classification of microplastics in water

Jaden Tolbert, Md Saiful Islam, and Pingshan Wang
Clemson University, SC 29634

**Abstract:** Micro- and nano-plastic particles (MPs/NPs) are ubiquitous environmental contaminants whose increasing abundance and potential health impacts have created an urgent need for rapid, label-free detection methods. As particle size decreases to the low-micrometer range, conventional optical and spectroscopic techniques become increasingly challenging because of limited throughput and/or complex sample preparation. In this work, we present a machine learning (ML)-assisted radio-frequency (RF) dielectric spectroscopic cytometry (DiSC) platform for the label-free detection and classification of MPs. Eight types of 10 µm nominal-diameter MP particles suspended in deionized (DI) water were characterized at four frequencies spanning 0.2–9 GHz. The measured alterations in RF scattering parameters (S-parameters), referenced to the carrier medium, were used to train supervised ML models for material classification, including the identification of MPs in mixed samples and saline-water environments. For eight MP classes suspended in DI water, the proposed method achieved macro-average F1-score, precision, and recall values exceeding 0.71. Furthermore, PET classification performance was largely maintained in saline carrier media containing 3.3% and 6.6% sea salt. These results demonstrate the feasibility of ML-assisted RF DiSC for rapid, single-particle MP classification in aqueous environments. Future work will focus on improving classification performance through enhanced RF calibration, increased spectral coverage, larger training datasets, and validation using environmentally aged and biologically contaminated microplastics.

## I. Introduction

Microplastics (MPs) and nanoplastics (NPs), ranging from 1 µm to 5 mm and 1 nm to 1 µm [1], respectively, are ubiquitous and emerging environmental pollutants in all types of water [2], soil [3], air [4, 5], and food [6-10]. Also, MPs/NPs have been reported in every organ, including heart and brain, of healthy volunteers [11]. Recent efforts confirmed that MPs/NPs cause hematopoietic damage, lung injury [12, 13], and immune toxicity [14] as well as disorder of gut microbiota, metabolites, and cytokines [15]. It is recognized that MP/NP effects are determined by MP/NP size, material [5, 16], shape [17], size distribution, and concentration [18], among others. Their potential adverse health effects have emerged as a great concern [7, 19-21]. Recently, the EPA is beginning to formally assess if MPs should be listed as a drinking water contaminant. Therefore, comprehensive measurement of MPs/NPs is important [22].

Optical methods, including naked eye observations, fluorescence microscopy, and flow cytometry (FC) [23], are commonly used for MP/NP detection and identification. Among them, FC [24] and flow imaging are capable of detecting MPs down to ~ 0.5 µm and 2 µm, respectively. They can be combined to provide information, potentially in situ, on MP sizes, shapes, possible aggregation. But downstream chemical analysis is needed to determine the type of MP/NP materials.

A few types of spectroscopic and thermal degradation approaches have been used to determine MP/NP materials. Fourier transform infrared (FTIR) spectroscopy [25] or FTIR imaging-based methods have been popular due to a good tradeoff between analysis time demands and level of detail. But FTIR methods are limited to dry samples with MPs larger than 10 µm. Raman methods [26, 27] can detect particles as small as 1 µm, render information on MP/NP materials, particle count, size distribution and morphology, and be potentially field deployable. However, Raman methods require rigorous sample preparation, including the extraction of MPs from water onto membrane filters, because they are highly susceptible to interference from organic and inorganic contaminants and background fluorescence. Additionally, Raman measurement requires extremely long sample analysis time and can easily damage MPs/NPs. Thermal degradation methods are also used with subsequent gas-chromatography (GC) - mass spectrometry (MS) analysis, such as pyrolysis-gas chromatography time-of-flight MS (PY-GCToF) [28]. They enable simultaneous identification and quantification of different MPs/NPs in complex environments as a main method in many plastic-biological samples. However, this method is destructive with low and polymer-dependent sensitivity [29]. Their applications in biological matrices have been challenged by interference or nonspecific products [30]. Additionally, the analysis is very time-consuming and requires comprehensive preparation for environmental samples.

Other methods have also been explored. For instance, hyperspectral imaging (HIS) can be easily added to micro-spectroscopy systems (e.g., Raman), creating multimodal systems without compromising the performance of either technique. This allows application of co-localized micro-spectroscopy and data fusion, where HIS can rapidly screen and discriminate micro- or nanoparticles. Scanning electron microscopy (SEM) allows for the investigation of size, shape, crystallography, and other physical and chemical parameters of particle surfaces down to a size of a few hundred nanometers. Even so, the measurement process is time-consuming. Size exclusion chromatography is another option, but it has a low limit-of-detection (LOD) and cannot identify MP/NP materials [2].

Therefore, new techniques to rapidly measure MP/NP properties, especially for material determination, are needed. We reported that radio frequency (RF) spectroscopy measures molecular composition profiles (MCPs) in biological cells [31]. The conservation of MCP in members of a common cellular species and physiological state enables machine learning (ML) based cell classification [32]. For MPs, their MCPs are different due to the use of different polymers and additives, which likely exhibit polymer-dependent polarizability in radio frequency domain. The dependence arises from the combined effects of polarizable groups in the molecular chain. The polarization and energy loss directly correspond to the dielectric properties of materials over frequency. Thus, the RF spectra of MPs may enable the differentiation of MP types. A few recent reports, such as [33, 34], alluded to such microwave spectra based classification, but no MP classifications or RF frequency markers were presented. Furthermore, no complete spectral results on single MPs for RF MP classification have been reported. In this work, we show reproducible measurement of MPs with an RF dielectric spectroscopic cytometer (RF DiSC) and RF spectra-based ML classification of MPs.

## II. Considerations of RF DiSC design and ML development

Fig. 1(a) shows the diagram of the RF DiSC system with an assembled sensing device (Figs. 1(b)) for reproducible MP measurement at single MP resolution (Fig. 1(e)). The basic operation principle has been reported elsewhere, e.g., in [35]. Specific considerations for MP measurement are summarized in section A below. Corresponding ML development approach for MP classification, illustrated in Fig. 1(f), is described in section B.

### A. RF DiSC measurement system

The measurement setup (Fig. 1(a)) consists of a vector network analyzer (VNA) (up to 9 GHz) and a microstrip based sensing device (Figs. 1(b) and (d)), which enable reproducible measurements and measurement of single MPs (Figs. 1(c) and (e)). The designed channel height of the microstrip-microfluidic device is 16 µm (*H* in Fig. 1(d)), appropriate for measuring 10 µm diameter MPs. A microscope and a video camera were used to record the MPs as they were measured. The video feed enabled estimation of MP size using average pixel diameter as a feature for classification. The particle motion was manipulated by using a push-pull pressure controller connected to one end of the device's tubing.

The microfluidic channel walls are formed with SU8 (Fig. 1(d)), which can absorb water facilitated by high pressure and long operation time [36]. Therefore, thorough rinsing/cleaning and drying of the sensing device was part of the measurement protocol established to achieve measurement consistency.

Scattering parameters (*S*-parameters) are recorded by the VNA. Measured $S_{ij}(f),\ i,j=1,2$ at frequency $f$ provides four *S*-parameters while each complex parameter has its own magnitude $S_{ijm}$ and phase angle $S_{ija}$. The VNA measures only two parameters at a time, so only $S_{11}(f)$ and $S_{21}(f)$ were used in this work. The *S*-parameter difference between water and MPs, $\Delta S(f)$ (e.g., $\Delta S_{11m}(f)$ in Fig. 1(e)), corresponds to their permittivity difference, $\varepsilon_{mp}(f)$- $\varepsilon_{water}(f)=\Delta\varepsilon(f)$. The RF spectra, i.e., $\varepsilon_{mp}(f)$, are the basis for MP classification. Nevertheless, in this work, we directly use $\Delta S\ (f)$ instead of $\Delta\varepsilon(f)$. Part of the reason is that the small $\Delta S\ (f)$ should stay unchanged even when the *S*-parameter baseline fluctuates (Fig. 1(e)) while water's properties are unchanged and $\Delta S$ is small. However, when $\varepsilon_{water}(f)$ or the sensing section in Fig. 1(b) changes, $\Delta S\ (f)$ will be affected even when $\varepsilon_{mp}(f)$ is not. As a result, the measurement system in Fig. 1(a) should be calibrated to extract the intrinsic $\varepsilon_{mp}(f)$. Appropriate calibration methods, such as the single-connection approach [37], are available for such operations. However, the method in [37] requires accurate simulation of the sensing section in Fig. 1(b) to the 3rd or even 4th digit of the *S*-parameters. We currently lack the computing power to achieve the accuracy needed.

The broadband signal of the measurement setup, Fig. 1(c), was recorded before and after MP measurement for confirmation of device consistency. A few factors were found to influence the broadband signal, affecting reproducibility of measurement results, including insufficient device rinsing after using cleaning solutions, insufficient time in vacuum before measurement, and particles sticking underneath the sensing line. When the broadband signals were confirmed to not have significant differences, the obtained MP data are used for ML training and testing. If an MP was stuck to the sensing line, did not fully pass through the sensing zone, or crossed the sensing zone with other particles (e.g., Fig. 1(d)), the signal was disregarded in further procedures.

In addition to $S(f)$ (e.g., Fig. 1(e)), MP images (Fig. 1(d)) are simultaneously recorded. Due to wide MP size distributions (Appendix, Fig. A1), imaging is expected to be an important MP sensing attribute for ML development.

### B. Data preprocessing and ML consideration

Both $\Delta S\ (f)$ and MP image data are analyzed for ML model training and testing. The obtained eight $\Delta S_{ij}(f)$ components are not independent from each other since they are fundamentally determined by $\varepsilon_{water}(f)$, $\varepsilon_{mp}(f)$, MP size, and measurement setup. Also, the RF DiSC system does not yield uniform $S_{ij}(f)$ responses among all the eight components, i.e., $\Delta S_{i,j}(f)$ has different signal-to-noise ratio (SNR) at each measurement frequency $f$. Thus, many of the factors that affect broadband $S$-parameters ($S_{ij}(f)$), such as reconditioning the sensing device in Fig. 1(b), are expected to induce measurement variabilities in $\Delta S_{ij}(f)$. As a result, ML development and MP classification performance will be affected.

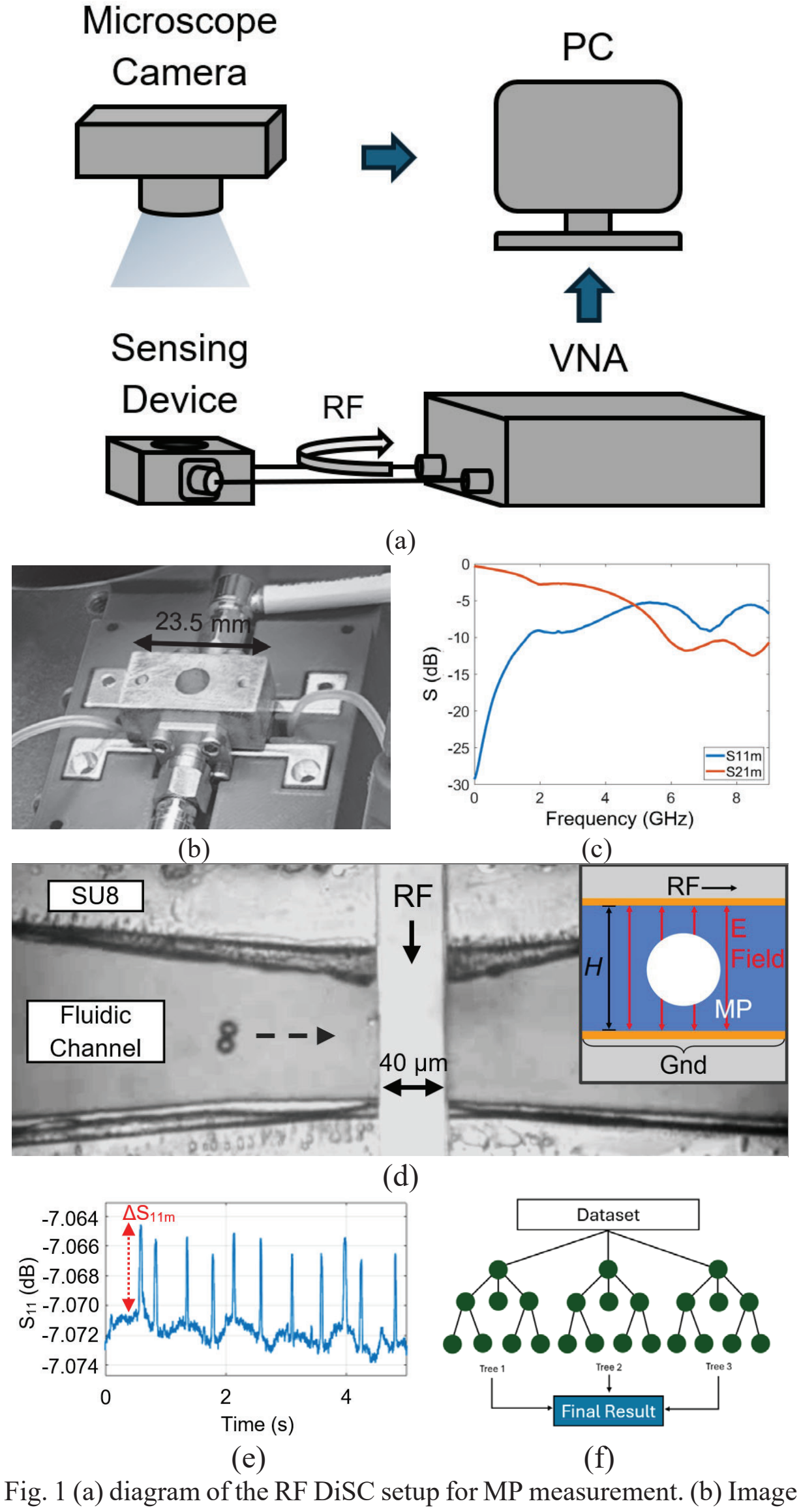

Fig. 1 (a) diagram of the RF DiSC setup for MP measurement. (b) Image of an assembled sensing device. (c) Typical broadband $S_{11m}$ and $S_{21m}$. (d) Two plastic particles approach the sensing zone with $H$=16 μm. (Top right: 2D cut-out/slice view of an MP in the RF sensing zone.) (e) Raw $S_{11m}$ measurement of an MP at $f$ = 9 GHz. (f) An ML architecture for development of an MP classification model.

Based on median absolute deviation (MAD), an algorithm was developed to process the raw S-parameters and calculate $\Delta S$ through the following steps: remove high-frequency noise by applying a light smoothing filter; detect events (i.e., MP presence) with a selected ($\Delta S_{11a}$) threshold value (24 in Fig. 2(a)); determine the left/right (start/end) points of the MP event by walking down the slopes of the signal until it flattens relative to the baseline (L/R in Fig. 2(a), zoom-in); create a mask to remove the MP event signal between points L and R; recalculate the estimated signal baseline after connecting L and R; compute $\Delta S$, i.e., the difference between the baseline and the signal peak, for ML classification.

Measured $\Delta S_{11a}$ and $\Delta S_{21a}$ showed the highest SNR at all measured frequencies. Thus, either was selected as the parameter to set a threshold to detect MP events. A relatively low threshold of ~3-times the median deviation of the selected $S$-parameter was found to be sufficient to consistently identify MP signal events. To reduce the chances of false detections, threshold values of 12 (0.2, 0.775, and 3.75 GHz) and 24 (9 GHz), Fig. 2(a), were used. The SNR for 9 GHz is higher; thus, a larger threshold is used. The results were manually reviewed after processing to ensure accurate event detection.

There are cases, such as Fig. 2(b) showing a particle reversing direction before exiting the sensing zone, where the signal baseline estimation does not match the actual baseline. The MP events are discarded after review. The number of measured MP events for each of the four selected frequencies is often inconsistent. Consequently, single MP events (*ΔS)* from the same MP at other frequencies were reused for compilation of additional training samples so that all the available MP events from all frequencies were utilized, Fig. 2(c). This data extension process generated additional unique samples for ML models to train on and was not implemented during compilation of the test data sets.

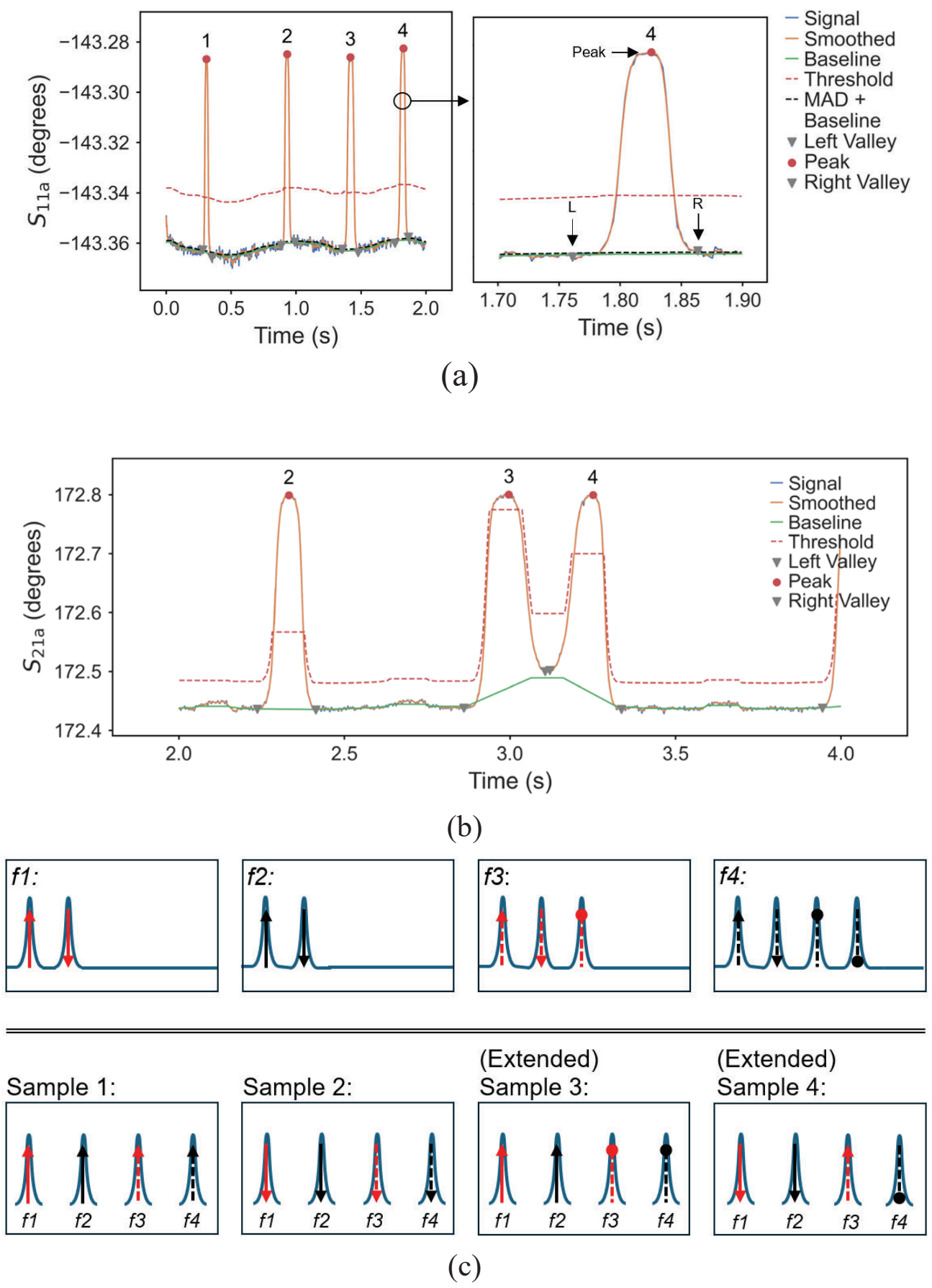

Fig. 2 (a) $S_{11a}$-based event detection with a line showing the baseline + MAD (i.e. threshold= 1) and event detection thresholds. A zoomed in view of a processed MP measurement event with marked peak, left, and right points. (b) An example of baseline tracking errors. (c) An illustration of *ΔS* sample curation for ML training. Signals are reused in order of occurrence at frequencies with less measurements.

Imaging is used to estimate MP size by a combination of FFmpeg filtering and OpenCV image tracking (Figs. 3(a) and (b) below).

Using MP *ΔS* and size as input features, principal component analysis (PCA) was investigated for dimensionality reduction and effective MP discrimination [38-39]. Bayesian model search [40] using a tree of Parzen algorithm [41] was performed to simultaneously select and optimize the model architecture best suited for classifying MPs by type during model training. The model search included ML methods such as support vector machines (SVM) [42-46] and tree-based ensembles such as random forests [47] and XGBoost [48]. Random forest models consistently produced the best results. SVM models exhibited significantly lower precision and recall, primarily because of difficulties in distinguishing PE from PET, PVC from ABS, and PS from PC. XGBoost models yielded results similar to those of random forests after extensive parameter tuning but did not reach the performance of the best-performing random forest models.

Group k-fold cross-validation was used to prevent data leakage from the same particle and provide more realistic estimates of model performance during training. Data samples originating from the same MP particle were grouped such that they remained within the same fold, and validation scores were determined by a majority vote across the samples within each MP particle group. Additionally, the folds were shuffled, and cross-validated three times during training to obtain more reliable estimates of the mean and standard deviation used for parameter optimization. Penalized optimization objective functions were implemented to achieve structural risk minimization.

## III. RF measurement of MPs and ML classification

Among common microplastics [49], polyethylene (PE), polypropylene (PP), polystyrene (PS), Polyethylene Terephthalate (PET), Poly(d,l-lactic acid) (PLA), Polyvinyl Chloride (PVC), Acrylonitrile Butadiene Styrene (ABS), and Polycarbonate (PC) were obtained from Echo Biosystems for this study. These MPs had a nominal size of 10 µm, with each type having different size distributions ranging from ~7-15 µm, shown in Fig. A1 (Appendix). PE and PET particles were PS based, having PS content less than 25%. Polyethylene (PE) and polypropylene (PP) are by far the most prevalent microplastics manufactured and found in the environment, followed by polystyrene (PS). Fig. A1 also lists the available dielectric properties of the plastic materials at given frequencies.

**A. MP RF spectral data collection:** The MPs were suspended in DI water and measured at a concentration of approximately $2\times10^6$ particles per mL to enable quick consecutive measurements while maintaining sufficient distance between particles to obtain single MP data, summarized in Table I. Fig. A2 shows typical $S_{11}$ and $S_{21}$. During RF data collection, the motion of MPs was monitored using video feed from the microscope and manipulated using the pressure controller. The monochrome microscope camera was 1.7 MP with 1 µm pixels and provided 96 fps video, sufficient for tracking particle motion.

To determine measurement frequencies $f_m, m = 1, 2, \ldots, n$ for $S_{ij}(f_m)$ , preliminary measurements were conducted on a limited set of plastic particle types (PS, PE, etc.). Based on these initial results and observations, 0.2, 0.775, 3.75, and 9 GHz (i.e., $n=4$) were selected due to comparatively large inter-type $\Delta S$ differences at these frequencies along with having a wide range of frequency coverage.

The measurement efforts lasted for several weeks. During the process, channel clogging events occurred and high pressure flushing and chemical-based de-clogging operations were performed. As discussed above, the operations may deteriorate measurement reproducibility. Three groups of data obtained.

The first group consisted of individual MPs suspended in DI water. In accordance with our measurement protocol, each set listed in Table I represents all MP measurements performed on a single day. To minimize potential cross-contamination, only one type of MP was measured per day. Each MP was measured repeatedly 30–100 times at each of the four frequencies. At the end of each day (i.e., each set), the device (Fig. 1(b)) was flushed and cleaned. Two measurement sets were targeted, with the training dataset designed to be twice the size of the test dataset. The measurement order of MP sets were shuffled to avoid the same MPs training and test set being measured consecutively for all types. However, for half of the types (PLA, PC, ABS, and PVC) the training and test did consist of consecutive measurement sets. Measuring the training and test sets for an MP consecutively risks the model learning measurement setup similarities to classify MPs, resulting in unrealistically high prediction precision. This is less of a concern when there is sufficiently low measurement variability between sets.

TABLE I: ML DATASET SUMMARY

| Data type | Particle type | Sets | Np | Suspension medium |
|---|---|---|---|---|
| Single MP training (Ns=8187) | ABS | 3 | 22 | DI water |
| | PC | 3 | 23 | DI water |
| | PE | 2 | 15 | DI water |
| | PET | 2 | 18 | DI water |
| | PLA | 1 | 18 | DI water |
| | PP | 2 | 22 | DI water |
| | PS | 3 | 16 | DI water |
| | PVC | 2 | 25 | DI water |
| Single MP test (Ns=3229) | ABS | 1 | 8 | DI water |
| | PC | 1 | 15 | DI water |
| | PE | 1 | 5 | DI water |
| | PET | 1 | 4 | DI water |
| | PLA | 1 | 11 | DI water |
| | PP | 1 | 6 | DI water |
| | PS | 1 | 3 | DI water |
| | PVC | 1 | 8 | DI water |
| Salted single MP test (Ns=269) | PET | 1 | 4 | DI water |
| | PET | | 3 | 3.3% sea salt |
| | PET | | 3 | 6.6% sea salt |
| Mixture training (Ns=890) | PE | 2 | 12 | DI water |
| | PP | 4 | 29 | DI water |
| | PS | 4 | 22 | DI water |
| Mixture test 1 (Ns=604) | 0.582 PE, 0.183 PP, 0.235 PS | 1 | 49 | DI water |
| Mixture test 2 (Ns=408) | 0.290 PE, 0.408 PP, 0.302 PS | 1 | 40 | DI water |
| Mixture test 3 (Ns=365) | 0.237 PE, 0.149 PP, 0.614 PS | 1 | 40 | DI water |

Np= number of particles, Ns = number of samples

Among the four frequencies, 9 GHz experienced worse baseline drift. So corresponding measurement time was increased to account for the likely increased rate of signal exclusion due to improper baseline tracking in signal processing.

The second group has sea salt added in DI water at a weight concentration of 0%, 3.3%, and 6.6%, mimicking conductivity fluctuations found in environmental water. PET particles were measured to assess the impacts of suspension medium on RF spectral measurement and ML classification performance.

The third is MP mixtures. PE, PP, and PS are selected because they are among the most produced plastic polymers. Instead of using RF data in the first group, separate MP training sets were obtained since the RF sensing device in Fig. 1(b) was altered significantly after the first two groups of measurements due to severe channel blockage and deep cleaning. High pressure and cleaning chemicals, including potassium hydroxide and Tween 20, were used. Broadband RF measurements verified permanent device alteration. Three test sets were measured with different MP mixture concentrations. Measurement data from two or more coincidental particles were ignored (e.g., Fig. 1(d)), possibly leading to some variation between the expected and measured concentration ratios. PE has a lower glass transition temperature, making particles more likely to stick together and stick to device channel surfaces.

**B. MP size estimation:** An image of the video feed without a particle present was used to improve background subtraction, Fig. 3(a), and ensure consistent particle bounding boxes during tracking (Fig. 3(b)). The average particle width ($x$), calculated from all video frames where the particle was fully visible, is used as the size feature for ML model development. In general, the variation in the particle size estimations was less than 1 pixel (out of 8-18 pixels). Manual pixel counting further confirmed the consistency of the size estimates across independent measurement sessions.

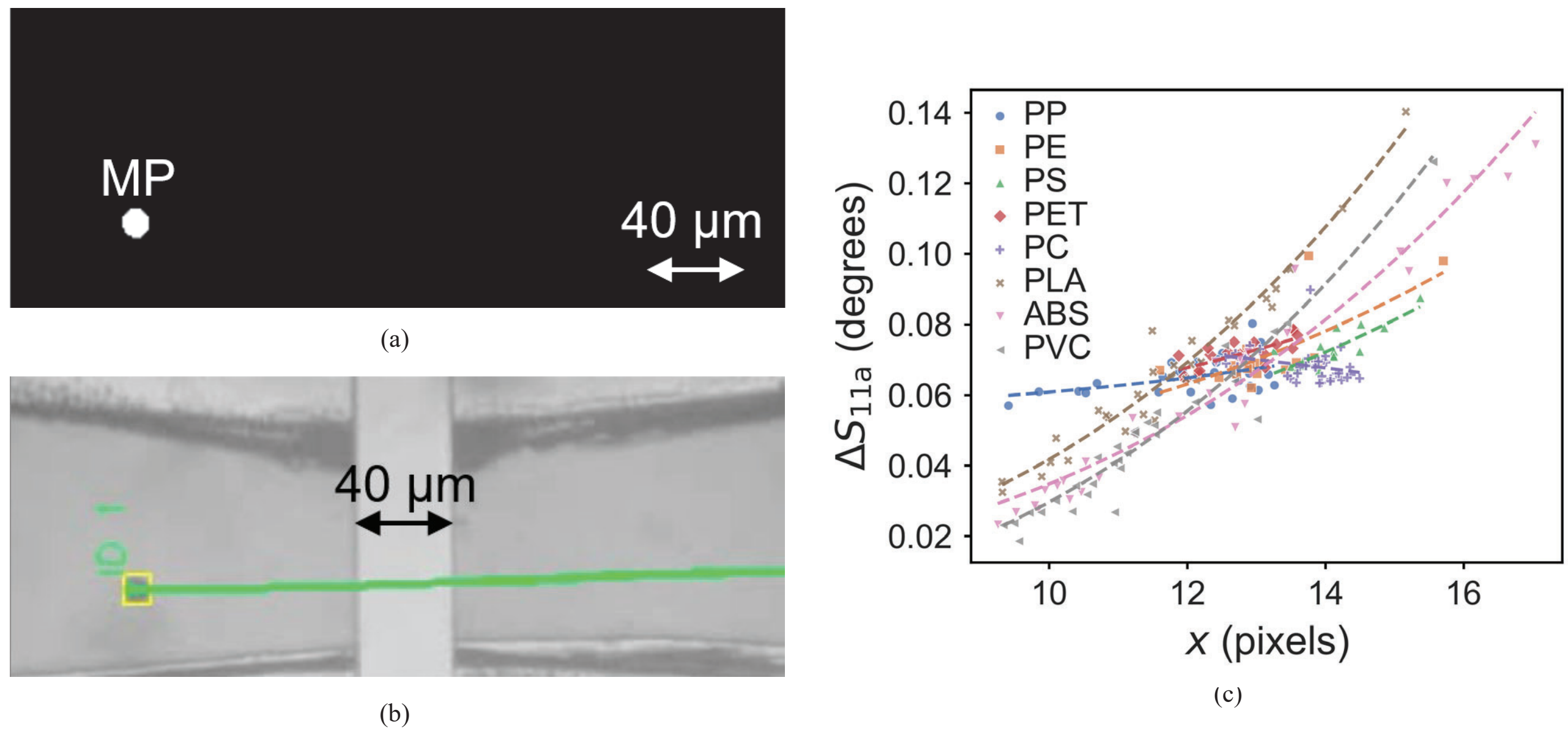

Fig. 3 (a) Video tracking: Background subtraction processing mask. (b) Video tracking: MP bounding box and path trace. The microstrip line width, 40 μm, was 42 pixels (c) Size estimation ($x^3$ fit) vs $\Delta S_{11a}$ at $f$= 0.775 GHz using the average signal from each particle.

The RF response of MP particles is expected to depend linearly on particle volume. Thus, a cubic relationship between particle width, $x$, and signal strength is expected. Fig. 3(c) presents the best-fit relationship between $x^3$ and $\Delta S_{11a}$. Particle types with relatively narrow size distributions exhibit weaker correlations, with $R^2$ values below 0.75. In contrast, ABS, PLA, and PVC particles have the widest observed size distributions and the strongest correlations, with $R^2$ values of approximately 0.95, 0.96, 0.93, respectively, when particle signals are averaged, and approximately 0.93, 0.97, 0.9, respectively, when individual particles are used at 0.775 GHz. The $R^2$ values are frequency dependent, with 9 GHz having the lowest values. Particle size estimation errors and SNRs are likely the primary factors limiting these correlations and both are expected to result in reduced ML classification accuracy.

**C. MP classification**: The $S$-parameter scatterplots of different MPs overlap at each of the measured frequencies, such as at 3.75 GHz in Fig. A3, indicating single-frequency RF features lack the information needed for classification of MP particles.

A two-component PCA was performed with and without MP size as a feature. Both analyses exhibited similar overall grouping patterns and substantial overlap among all particle types, although MP size inclusion yields slightly more separation between overlapping groups when ABS, PLA, and PVC are excluded, Fig. A4. PVC, PLA, and ABS account for much of the variance and are broadly distributed along the dominant principal component, with their groupings largely associated with MP size/signal level. For particles of similar sizes, the signals from different particle types were similar with differences close to the measurement noise, Fig. 4(a) and (b). Thus, the particle types are not trivially distinguishable through PCA analysis. This indicates that simple clustering or linear methods are less appropriate for classification. Difficult separation of the MP classes into groups leads to a higher risk of overfitting during training.

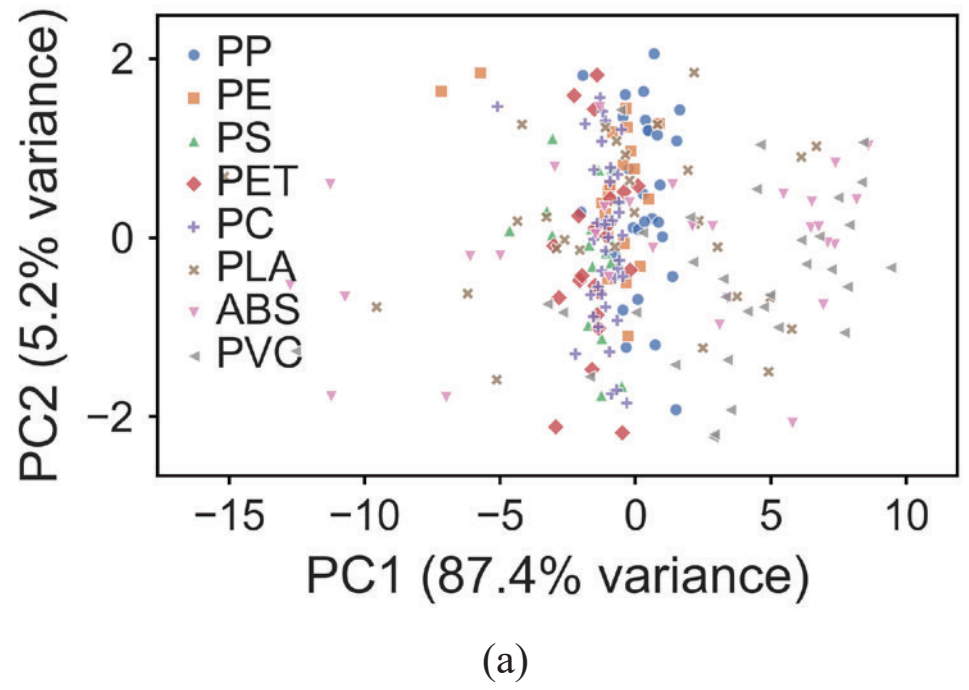

(a)

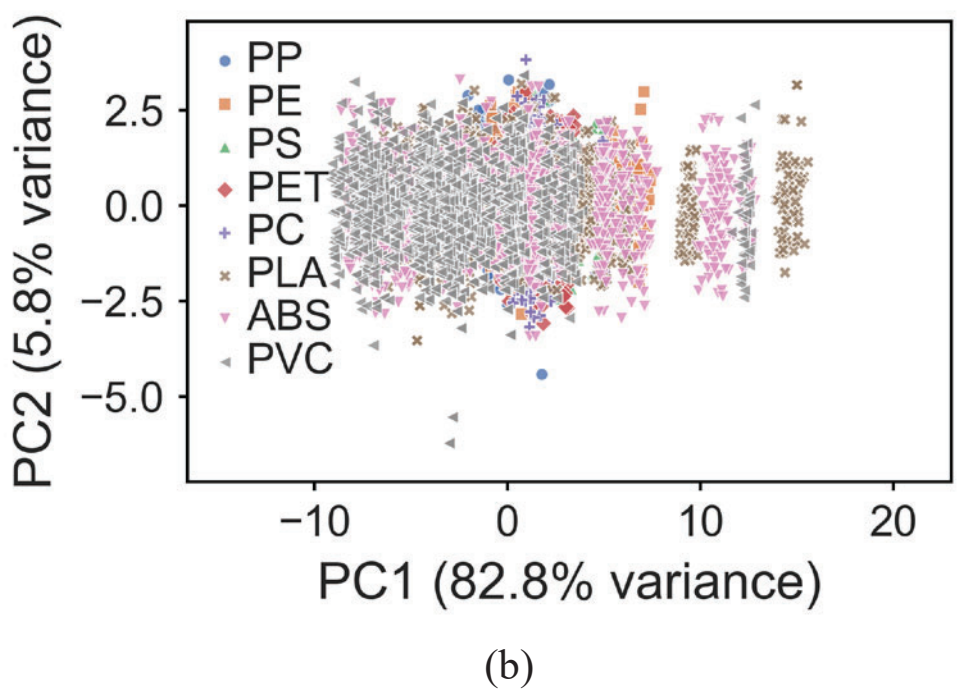

(b)

Fig. 4 A 2-component PCA using the single MP training and test datasets (a) Averaged signals per particle explaining 92.6% of the variance. (b) All individual samples explaining 88.6% of the variance

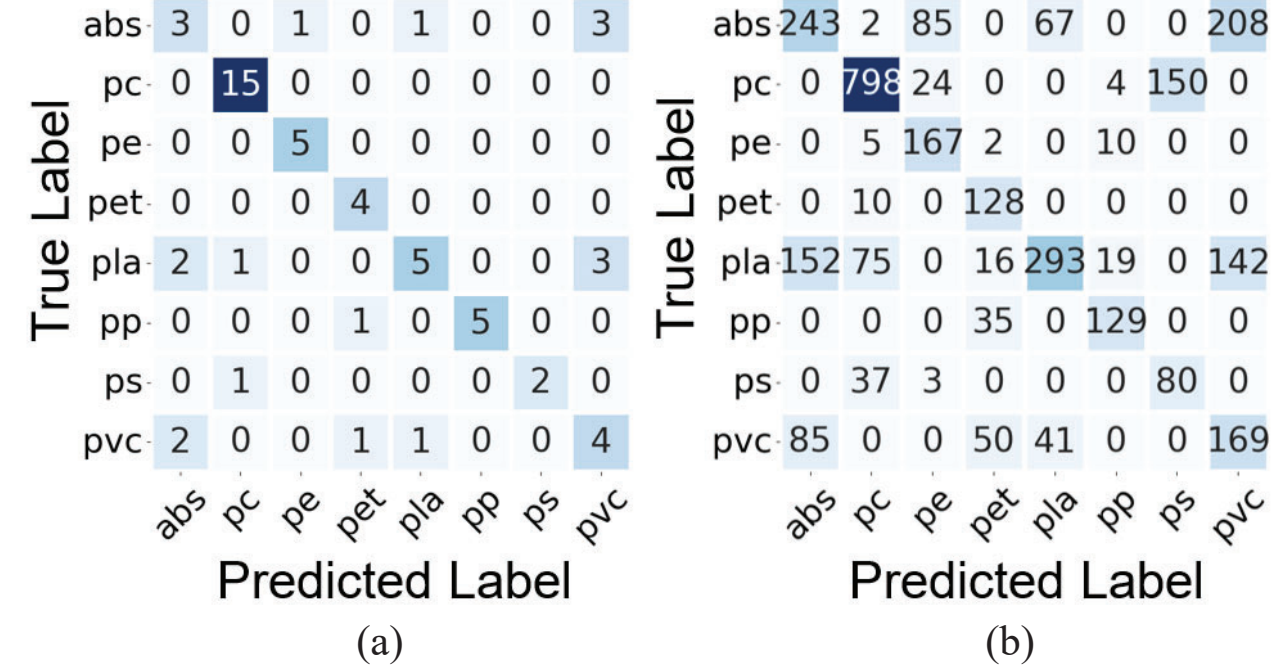

Fig. 5 (a) Majority vote per particle test results (Train/Test F1 macro: 0.73/0.72). (b) Individual sample test results (Train/Test F1 macro: 0.70/0.61)

A variety of ML models were trained to investigate the relationships between the selected features, training dataset size, and classification performance. Random forest classifiers achieved the best overall performance and were therefore selected, trained, and examined for MP classification.

The model training was performed through parameter optimization and a penalized macro recall scoring metric then evaluated through confusion matrices (Fig. 5), receiver operating characteristic (ROC) curves (Fig. 6), and multi-class precision-recall curves (Fig. 7). The area under the ROC curve (ROC-AUC) and mean average precision (mAP) across all classes were used as quantitative measures of overall model performance. Both individual sample (Fig. 2(c)) predictions and the majority vote of sample predictions for each particle in the test set were evaluated. These metrics provide a comprehensive assessment of model performance across all decision thresholds and characterize the tradeoff between precision and recall. Confusion matrices were further used to evaluate both overall performance and class-specific prediction trends. Precision (the fraction of predicted samples that were correctly classified) and recall (the fraction of true samples that were correctly classified) were evaluated for each class. Macro-averaged metrics, which weigh each class equally, were used for overall performance evaluation because the relative prevalence of MP types in these datasets is unrelated to those expected in practical applications. Potential overfitting was assessed by comparing the macro F1-scores of the training and test datasets (Table II).

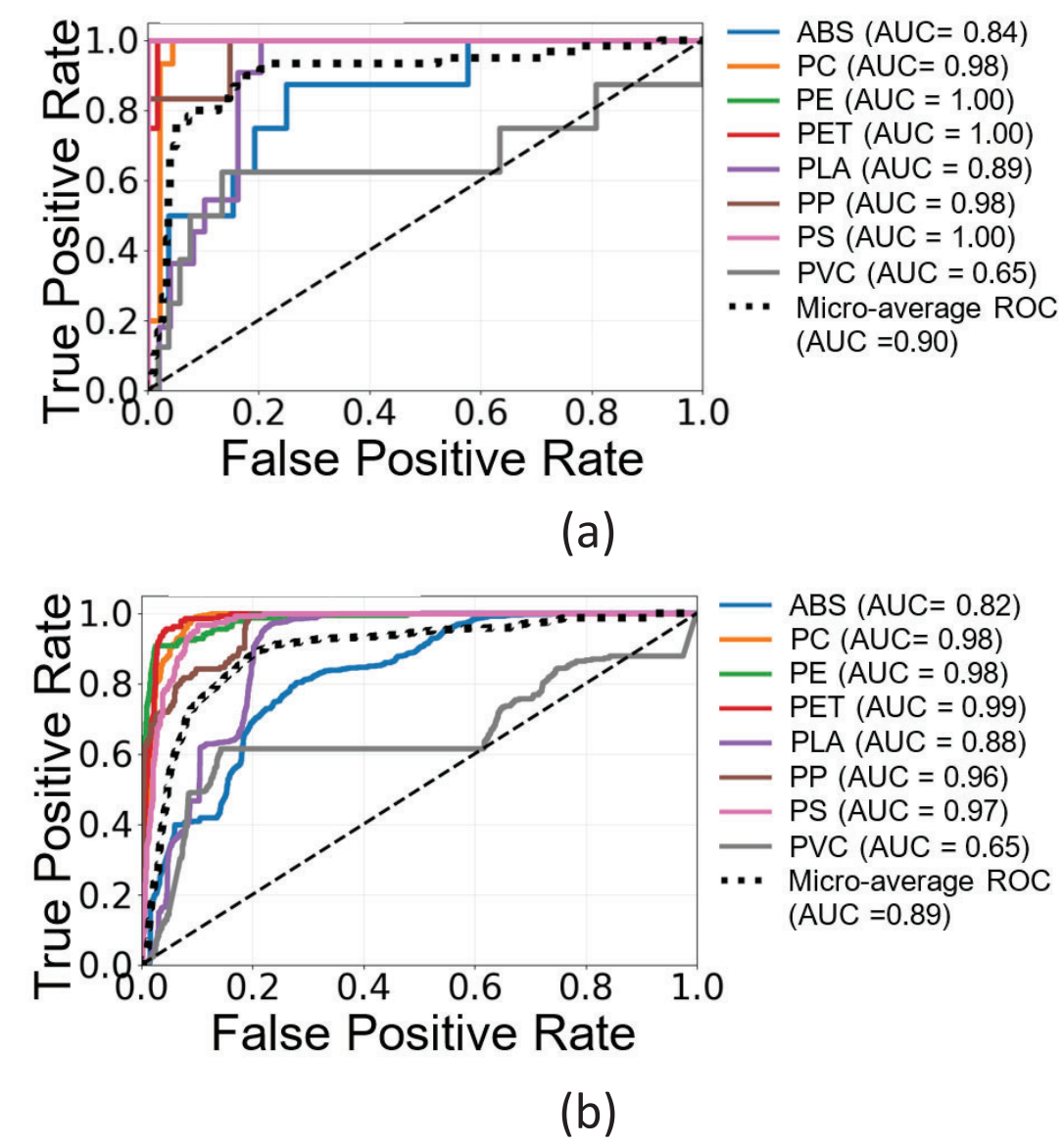

Fig. 6 ROC curves: (a) Majority vote per particle test results. (b) Individual sample test results.

Initial models were trained using all measured frequency and size features. During the training process, models tended to overfit very easily, causing a significant training-validation gap. High overfitting leads to unrealistically high training scores due to the model memorizing properties specific to the training set, rather than to the classes of MP types. A well generalized model will produce training scores similar to validation and test scores, meaning the model performs consistently, even for unseen data. In order to mitigate overfitting behavior, a penalized score using generalization and variance penalty coefficients on the validation scores were used to train the model during parameter optimization. It was found that a relatively high generalization penalty of 0.8 worked the best.

Feature importance analysis showed that $\Delta S_{21m}$ and $\Delta S_{21a}$ at 0.2 GHz and $\Delta S_{21m}$ at 0.775 GHz contributed negligibly to the model's predictions. These features also exhibited the lowest SNRs (approximately 1). Removing $\Delta S_{21m}$ at 0.2 and 0.775 GHz improved the macro F1-score and balanced accuracy by several percentage points. Subsequent removal of $\Delta S_{21a}$ at 0.2 GHz produced only marginal additional improvements, if any, and models tended to overfit easier.

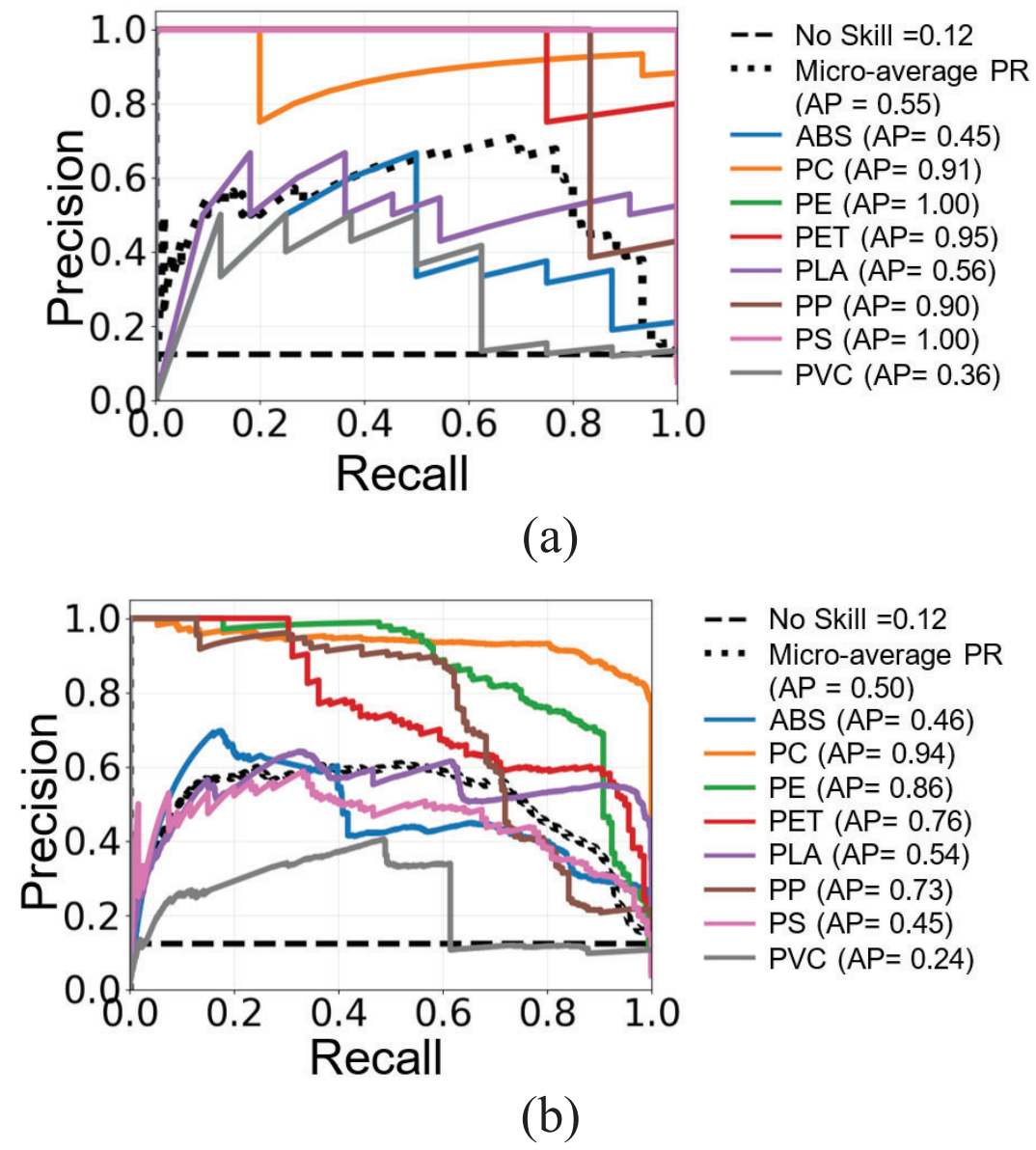

Fig. 7 PR curves: (a) Majority vote per particle test results (mAP: 0.77). (b) Individual sample test results (mAP: 0.62)

Additional feature engineering was performed by removing all other $\Delta S$ measurements from individual excitation frequencies to evaluate their importance to the performance of the model. Removing all other features associated with frequencies 0.2 and 0.775 GHz reduces the macro F1-scores to approximately 0.5. Further removal of 3.75 GHz features reduced the macro F1- score to below 0.5, demonstrating that information from all measured frequencies significantly contribute overall classification performance.

Particle size feature was consistently identified as the most important overall, and its removal resulted in a substantial decrease in classification performance. Nevertheless, models trained without the size feature still achieved ROC-AUC values greater than 0.8 for all particle classes except PVC, Fig. A5. For multi-class PR curves, PS classification was affected by size feature removal most severely, approaching random guess (0.12). Despite this reduction, average precision remained above random guess level for all remaining particle classes.

Additionally, normalization $\Delta S$ by dividing it with estimated MP volume while excluding or keeping the size/ width as a feature was attempted. These attempts either decreased performance or yielded results similar to including size as a feature, indicating that size normalization alone could not compensate for the loss of the explicit size feature.

TABLE II: CLASSIFICATION TABLE

| | **Precision** | **Recall** | **F1-score** | **support** |
|---|---|---|---|---|
| **ABS** | 0.51 | 0.40 | 0.45 | 605 |
| **PC** | 0.86 | 0.82 | 0.84 | 976 |
| **PE** | 0.60 | 0.91 | 0.72 | 184 |
| **PET** | 0.55 | 0.93 | 0.69 | 138 |
| **PLA** | 0.73 | 0.42 | 0.53 | 697 |
| **PP** | 0.80 | 0.79 | 0.79 | 164 |
| **PS** | 0.35 | 0.67 | 0.46 | 120 |
| **PVC** | 0.33 | 0.49 | 0.39 | 345 |
| ***Macro avg.*** | 0.59 | 0.68 | 0.61 | 3229 |
| ***Macro avg. (Majority-vote)*** | 0.74 | 0.73 | 0.72 | 60 |

Precision= $\frac{TP}{TP+FP}$, recall= $\frac{TP}{TP+FN}$, F1= $\frac{2*precision*recall}{precision+recall}$, (T= true, F= false, P= positive, N= negative). Macro averages give equal weight to each

Table III: Feature Importances

| Feature | Importance |
|---|---|
| *size* | 0.19 |
| *f3* ΔS11m | 0.09 |
| *f4* ΔS11a | 0.08 |
| *f4* ΔS21a | 0.08 |
| *f3* ΔS21a | 0.07 |
| *f2* ΔS11a | 0.07 |
| *f2* ΔS11m | 0.06 |
| *f1* ΔS11a | 0.06 |
| *f3* ΔS11a | 0.06 |
| *f4* ΔS11m | 0.06 |
| *f3* ΔS11m | 0.05 |
| *f2* ΔS21a | 0.05 |
| *f4* ΔS21m | 0.04 |
| *f1* ΔS11m | 0.04 |
| *f1* ΔS21a | 0.001 |

To evaluate the influence of the training dataset size, reduced training datasets were generated by uniformly removing samples from each particle class and set (measurement day), thereby minimizing dependence on any single measurement session. Even when only one-eighth of the original training data was retained, the model continued to perform better than random classification, with PVC exhibiting the lowest ROC-AUC (0.64). However, the mAP decreased to 0.56, indicating a substantial reduction in classification precision. Reducing the training dataset by one-half resulted in only a modest decline in performance. Models trained on half of the available data achieved ROC-AUC and mAP values of 0.89 and 0.77, respectively, compared with 0.90 and 0.77 for the best-performing model trained on the complete dataset.

Synthetic training data were also generated through interpolation of measured particle responses. The measured noise profiles and averaged particle signals from the full dataset were used to generate 40 synthetic samples per particle, and datasets containing 10, 50, and 120 particles per class were used for model training. Compared with the original training data, models trained on synthetic datasets exhibited weaker agreement between validation and test performance, indicating reduced generalization. The dataset containing 10 particles per class produced the lowest performance, whereas the 120-particle dataset performed slightly worse than the original experimental dataset. Nevertheless, models trained using the 120-particle synthetic dataset consistently classified all particle types better than random, achieving average precision values above 0.36 and ROC-AUC values above 0.60 for every class. Hence, synthetic data augmentation did not provide a significant improvement over training with the original experimental dataset.

A PCA preprocessing pipeline, having 3, 5, and 8 primary components, was also tested with the 5 component PCA-random forest model performing the best, achieving results shown in Fig. A6 and Table AII. The performance is close to the optimized random forest model trained on the complete experimental dataset with two low-importance features removed (Figs. 5-7 and Table II). For individual sample predictions, the model with 5 component PCA preprocessing resulted in ROC-AUC values exceeding 0.8 for all particle classes. Average precision remained above random guess for every class, resulting in an overall mAP of 0.52. Majority-vote classification produced slightly lower ROC-AUC values but yielded a higher overall mAP of 0.78. However, the 5 component PCA preprocessing still resulted in poorer precision and recall scores compared to the model trained without it.

The best performing model was an optimized random forest model trained on the complete experimental dataset with two low-importance features removed (Figs. 5-7 and Table II). The individual sample prediction precision and recall were 0.59 and 0.68, respectively. When using the majority vote of sample predictions from a particle, both precision and recall exceeded 0.72. Table AI details an example of the permutation importance on single particle test data. Table III summarizes the obtained Gini importances of the features for the model used to obtain the results in Figs. 5-7 and Table II. The model was able to consistently classify held-out PET measurements collected with and without added salt with high precision using both individual sample predictions and majority-vote particle predictions.

In group 2 of Table I, salt water and DI water presented different baseline *S*-parameters, along with different ΔS-parameter magnitudes. Salt water may also induce MP property changes or nonlinearly alter RF sensing fields. These effects are not incorporated into the ML model training and may contribute to a loss of MP classification accuracy. Nevertheless, held-out PET measurements acquired in media with and without sea salt content up to 6.6% consistently classified PET with high precision (Figs. A7(a)-(f)). Implementing a 5-component PCA preprocessing pipeline resulted in poor PET classification performance, indicating that the proposed method is robust to changes in salinity for PET particles while models with PCA preprocessing result in less robust predictions (Figs. A8(a)-(c)). However, it is unclear if changes in the surrounding medium may influence classification of certain particle types more strongly than others, meaning the effect of medium composition on predictions could be MP type dependent.

To test the ML model classification of MP particle mixtures, a model was trained using the Mixture training data in Table I and the same feature selection strategy discussed above, i.e., removing $\Delta S_{21m}$ at 0.2 and 0.775 GHz, to classify particles in the three mixture solution experiments. Although the training dataset for mixture experiments was smaller than that of the single-particle experiments, the predicted particle distributions generally followed the expected trends based on the MP mixture compositions, Fig. 8. However, the models systematically overpredicted PE in PE-rich mixtures and underpredicted PE when it was the minority component.

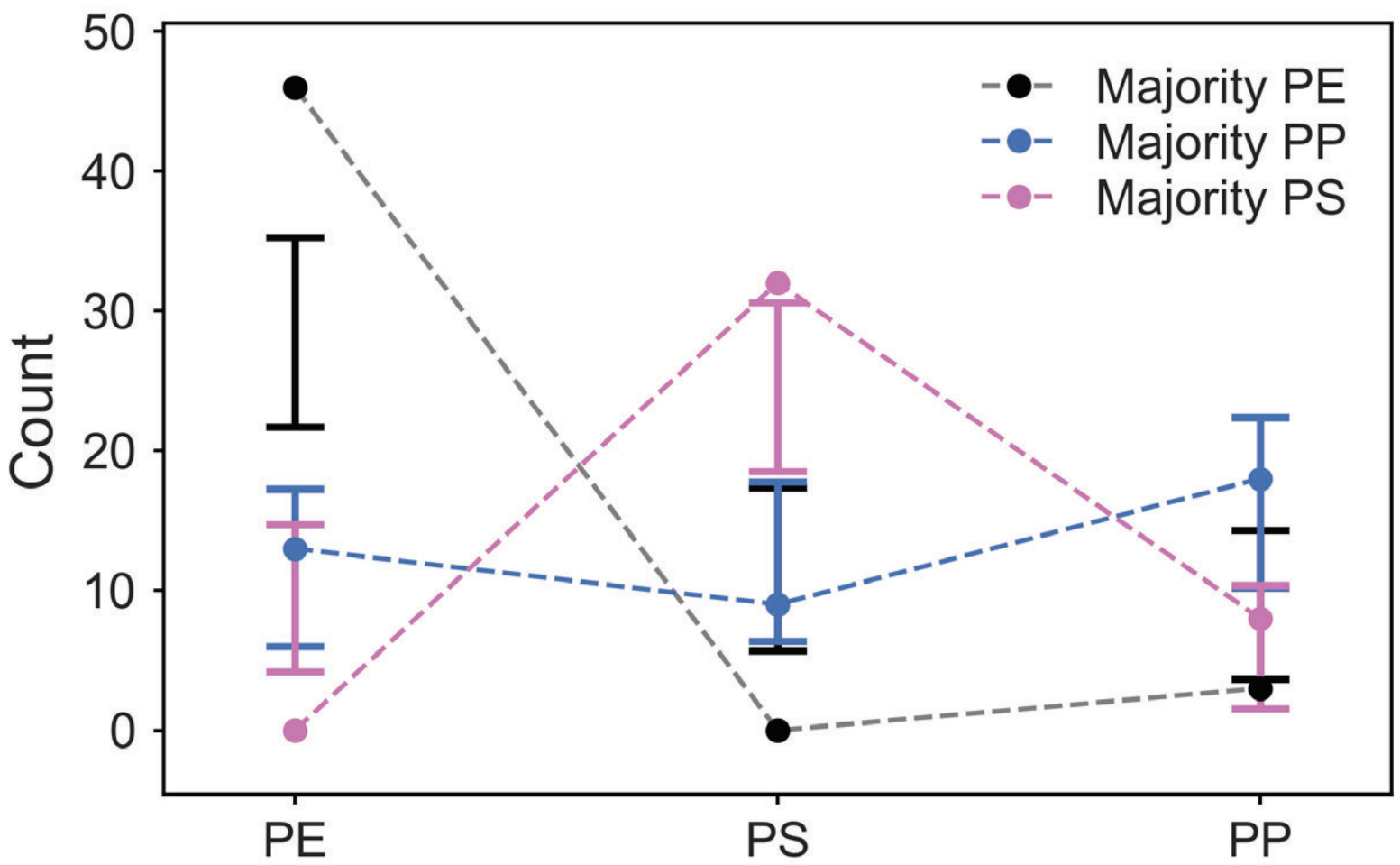

Fig. 8 Mixture MP test results: predicted particle count with expected margin of error (MOE). $MOE = Z \times \sqrt{\frac{p(1-p)}{n}}$ with Z=1.96 (95% confidence), n= total count of measured particles from mixture solution, p=expected MP portion of population ( $\frac{MP\ concentration}{total\ concentration}$ ).

## IV. Discussions

The worst-performing materials were ABS, PLA, and PVC. These MP particles also exhibited the greatest size variation. The number of intermediate frequencies were increased by adding measurements at 4.75, 6.25, 7, and 8 GHz for ABS and PLA, but was found to not improve classification performance. One possible explanation for the reduced performance is that these materials are dielectrically similar within the measured frequency range, which reduces separability. Another possibility is that the larger size variation of these MP types makes it difficult for the models to interpolate across relatively large gaps in the size distribution. However, this explanation is only partially supported, because synthetic data generation did not significantly improve performance. Nevertheless, classifying MPs down to small MP-type groups is still valuable.

While not the most important feature for every MP type, size was the most important feature overall. Therefore, bimodal sensing that combines RF DiSC and optical imaging, provides higher classification accuracy, consistent with our prior observations in yeast-cell classification [32]. Consequently, high-resolution imaging has the potential to further improve classification performance. However, size does not define MP material identity (i.e., $\varepsilon_{mp}(f)$ is size independent) even though it is a major factor in determining $\Delta S$, which is subsequently used for ML model development. MPs of different sizes have different surface-to-volume ratios, which may result in size-dependent surface polarization effects. Such effects often occur in the MHz frequency region. In the GHz range, the impact of surface polarization is generally considered negligible. Therefore, further investigation is of interest especially when normalization of $\Delta S$ by particle volume did not improve ML model performance.

RF sensing devices and environmental factors that do not alter MPs, such as small variations of the carrier liquid's temperature, pH, and conductivity, should not fundamentally affect RF-based classification of MP materials. However, these factors can influence the measured $\Delta S$ and, consequently, the performance of $\Delta S$-based ML classification models. Additionally, sensing device structure, carrier media property, and environmental factors discussed in section II will impact system performance due to significant changes of baseline $S$-parameters. Therefore, further work is needed to compensate for or eliminate the influence of these factors and recover the intrinsic dielectric properties of MPs, $\varepsilon_{mp}(f)$, through appropriate system calibration. As a result, RF-based MP sensing is expected to be standardized. Furthermore, by incorporating additional measurement frequencies across a wider frequency range and expanding the training dataset to include a substantially larger number of MP particles, we predict that the accuracy, robustness, and selectivity of RF-based MP sensing can be further improved.

The proposed RF-AI method is promising as a complement to various label-free, ML-assisted vibrational spectroscopy techniques. For instance, high-level fusion of attenuated total reflection Fourier transform infrared spectroscopy (ATR-FTIR) and Raman spectroscopy [50] achieved 99% classification accuracy and 99% recall for eight MP types, whereas individual Raman-only and ATR-FTIR-only models achieved 75% and 73% classification accuracy, respectively, using samples approximately 100 μm or larger and a 10 s Raman acquisition time. NIR hyperspectral imaging [51] reported a weight-average recall of 0.79 for 7 polymer classes compared with a macro-average recall of 0.71. Raman spectroscopy [52] achieved 0.99 recall and 99% classification accuracy of 10 common plastic types. In comparison, our RF sensing approach identified 10 μm MP particles across eight MP types, achieving majority-vote macro-average precision and recall of .74 and .73, respectively. Moreover, unlike the studies in [50-52], which primarily measured dry plastic samples, the proposed method directly measures MP particles suspended in water. Although its current classification performance is somewhat lower than that of optical spectroscopic techniques, the substantially smaller detectable particle size, scalability of the sensing structure, significantly faster measurement time, and capability for in-water measurements highlight the unique advantages of the proposed RF sensing approach. We anticipate that its performance will be greatly improved through enhanced RF system calibration, expanded frequency coverage, and larger training datasets.

## V. Conclusions

This work demonstrates the feasibility of machine learning (ML)-assisted RF dielectric flow cytometry for the rapid, label-free classification of microplastic (MP) particles suspended in water. The proposed RF sensing system, consisting of a microfluidic RF sensing device integrated with a vector network analyzer, enables single-particle dielectric measurements. Eight types of MPs with a nominal diameter of 10 μm were measured at four frequencies spanning 0.2–9 GHz. Using the measured RF scattering parameters (S-parameters), referenced to the carrier medium, a supervised ML model was developed for MP classification. For MPs suspended in deionized (DI) water, the model achieves macro-average F1-score, precision, and recall values exceeding 0.68 across the eight MP classes. The trained model was further evaluated using saline carrier media. For PET MPs suspended in 3.3% and 6.6% seawater, the classification performance was not significantly degraded, demonstrating the robustness of the proposed approach to moderate changes in medium conductivity.

Although the present classification performance remains below that of state-of-the-art optical spectroscopy methods, the proposed RF approach offers several unique advantages, including nondestructive direct measurement of single 10 μm particles suspended in water, millisecond-scale measurement time, and compatibility with microfluidic flow cytometry. Future work will focus on improving the dielectric spectroscopy and classification performance through enhanced DiSC sensitivity, reduced measurement variability, wider frequency coverage, more comprehensive RF calibration, and more expansive training datasets. In addition, the proposed method should be validated using environmentally weathered microplastics and particles recovered from biologically contaminated or other complex environmental matrices to further assess its practical applications.

## Acknowledgements

The authors acknowledge Cooper Taylor of Clemson University for helping with development of the MAD-based signal processing algorithm. The authors also acknowledge and thank Dr. Xiaoyong Brian Yuan and Dr. Carl Ehrett of Clemson University for their consultation concerning machine learning and data analysis.

This research was supported through funding by the USDA STTR Phase I Award (2025-51402-44885)

# Appendix

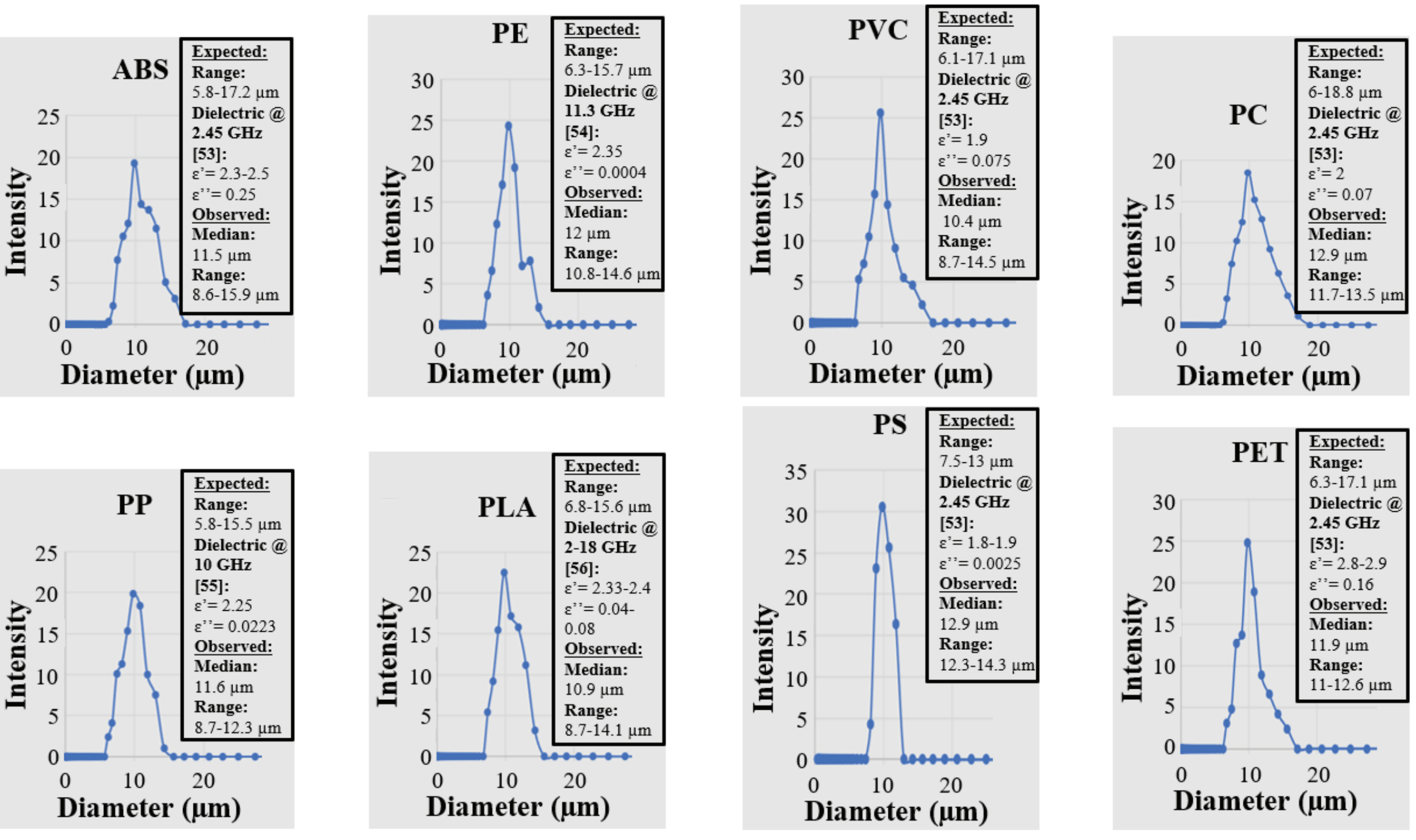

Fig. A1 Particle information based on size distribution charts from the manufacturer and published dielectric properties.

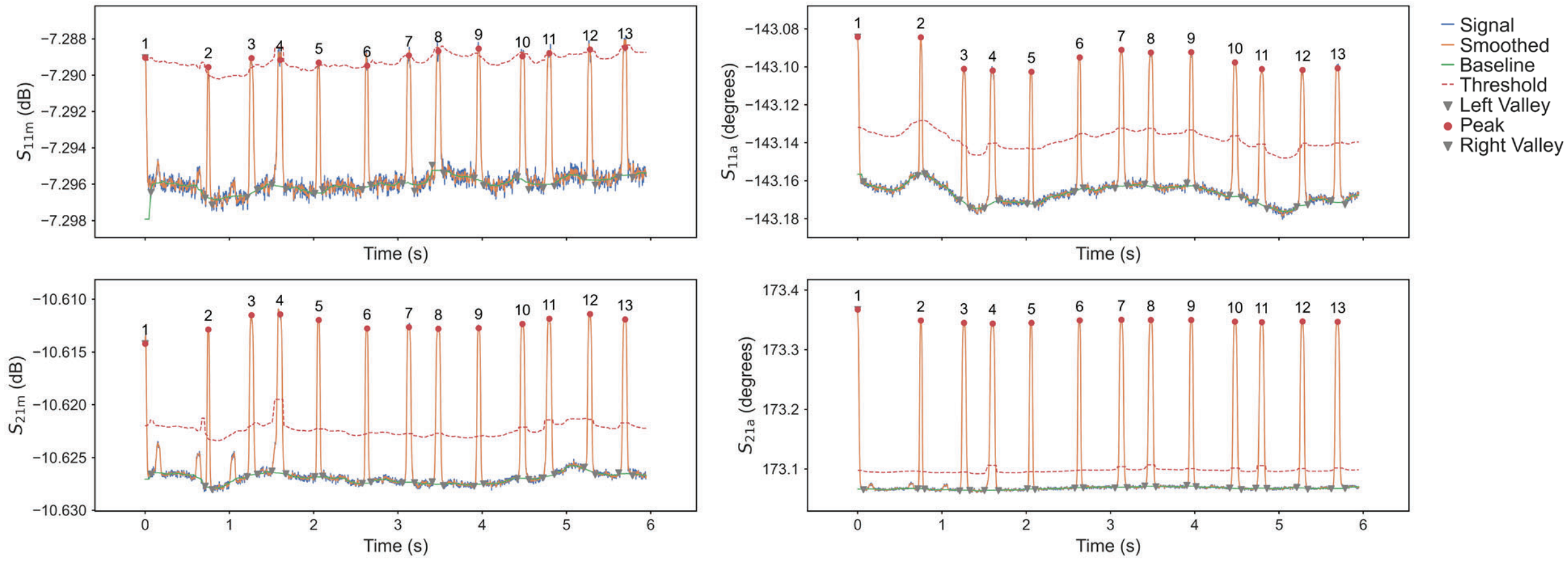

Fig. A2 Typical *S*-parameters at 9 GHz

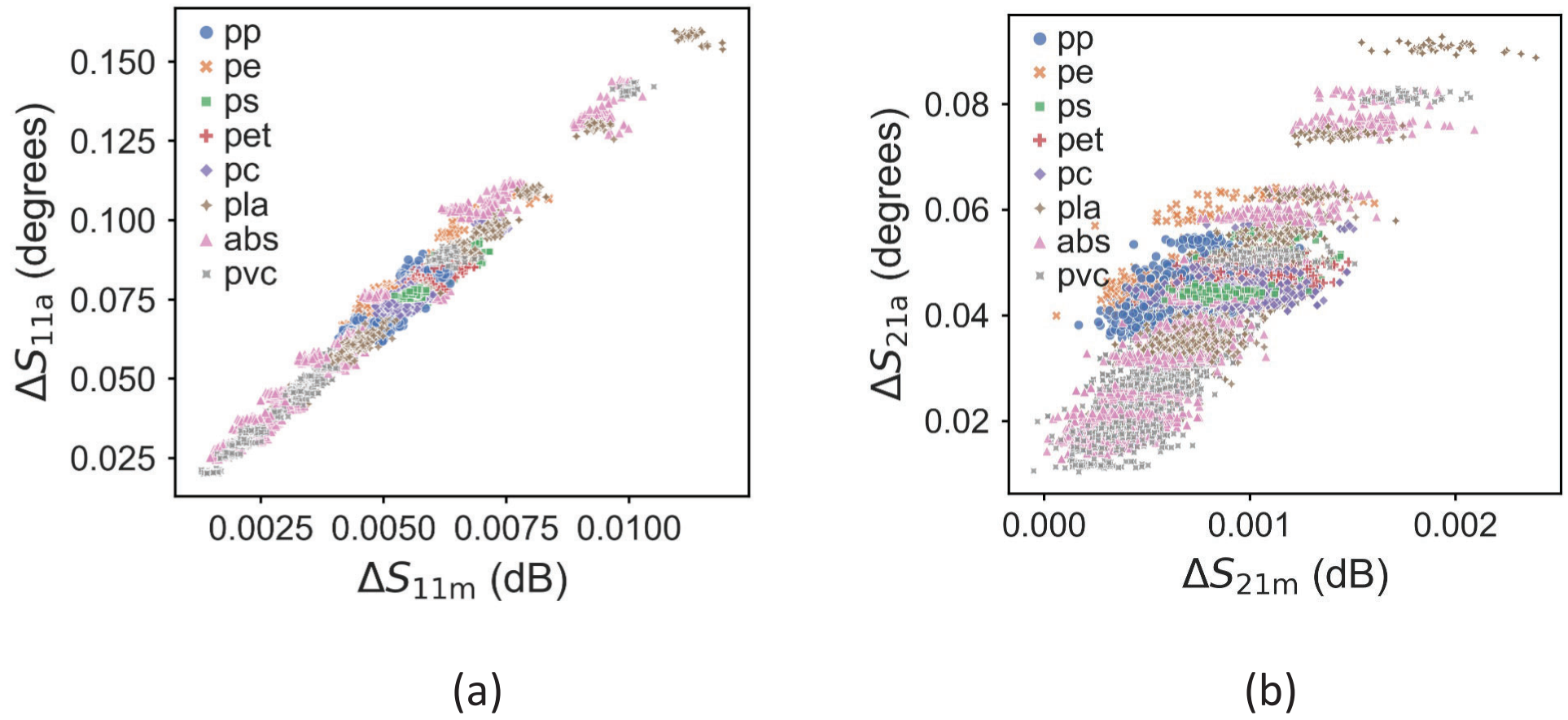

Fig. A3 $S_{11}$ (a) and $S_{21}$ (b) at 3.75 GHz

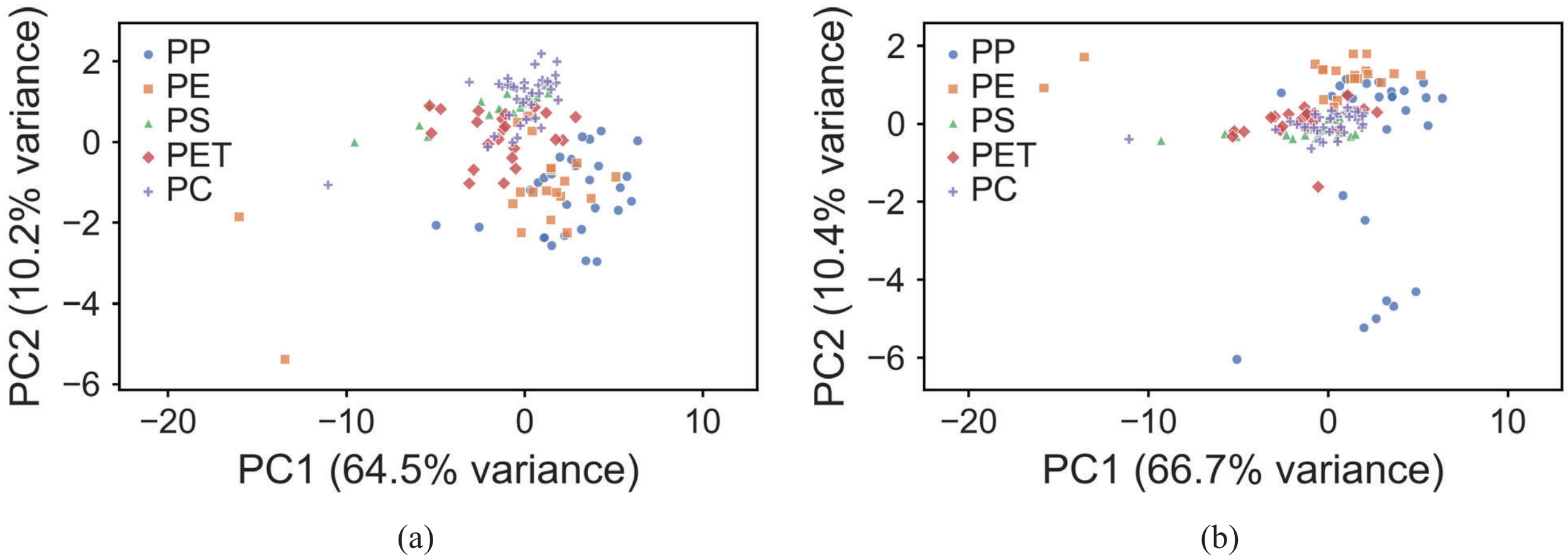

Fig. A4 2-component PCA with (a) and without (b) size when PCA, PLA, and ABS are excluded.

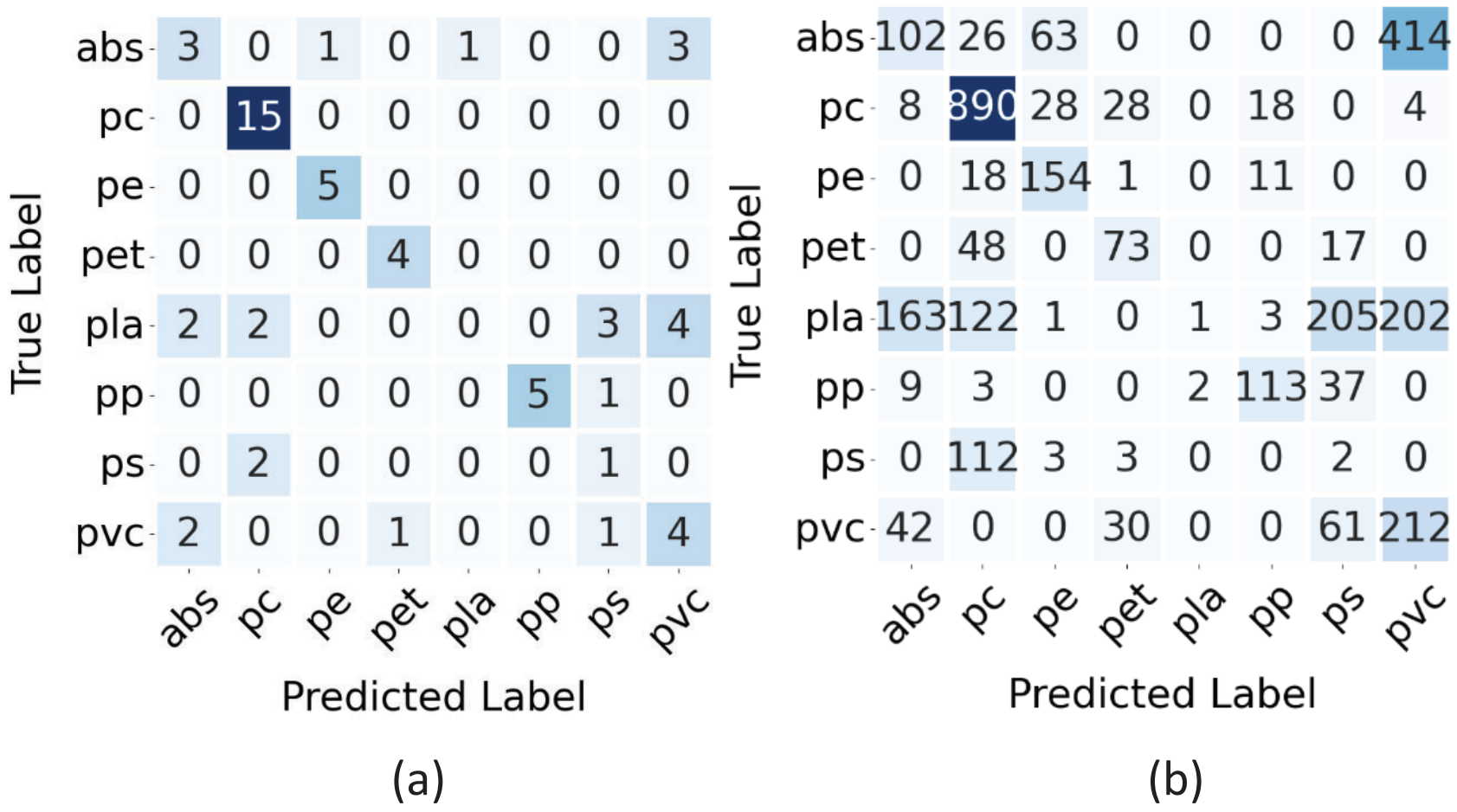

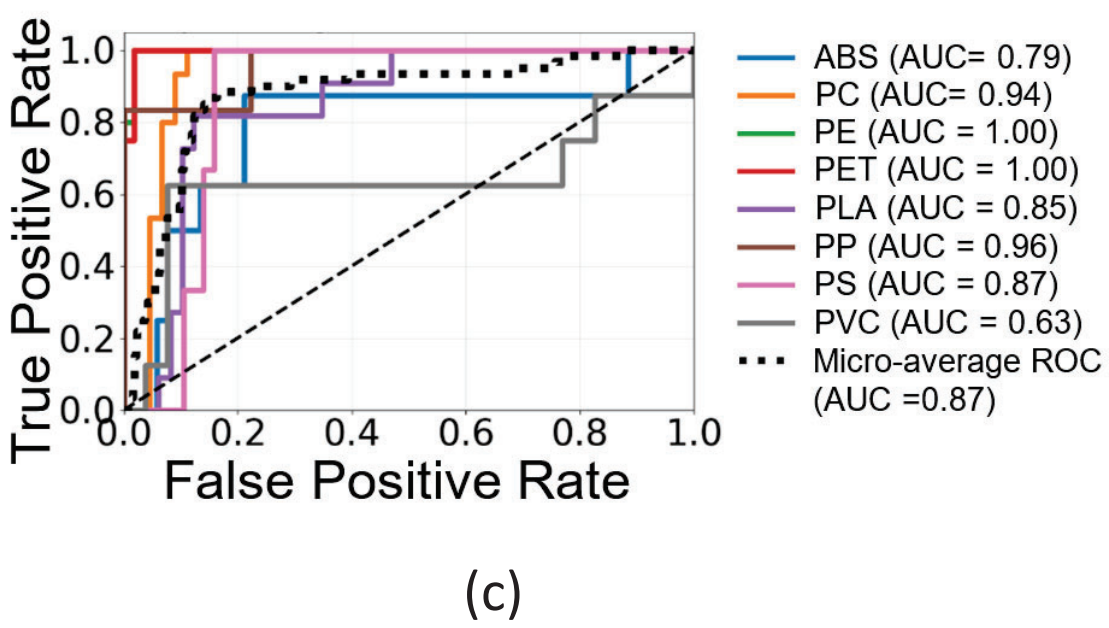

(c)

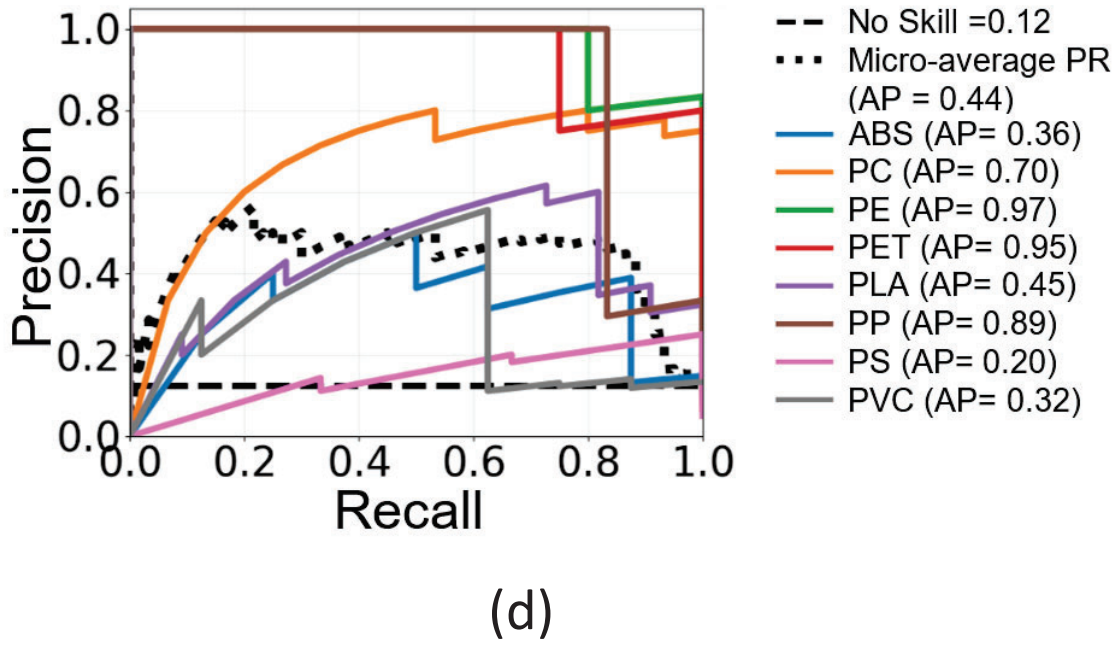

(d)

Fig. A5 Classification results without the size feature (mAP: .61): (a) majority vote per particle confusion matrix (b) Individual sample confusion matrix (c) majority vote ROC curves (d) majority vote PR curves

TABLE AI: SINGLE PARTICLE TEST PERMUTATION IMPORTANCES

| Feature | Mean Importance | Importance std |
|---|---|---|
| *size* | 0.173 | 0.007 |
| *f4* ΔS11a | 0.137 | 0.008 |
| *f3* ΔS11m | 0.080 | 0.003 |
| *f3* Δs21m | 0.028 | 0.004 |
| *f4* ΔS11m | 0.024 | 0.004 |
| *f2* Δs11a | 0.015 | 0.004 |
| *f2* ΔS11a | 0.014 | 0.003 |
| *f3* ΔS21a | 0.012 | 0.002 |
| *f4* ΔS21m | 0.011 | 0.005 |
| *f1* ΔS11a | 0.010 | 0.003 |
| *f4* ΔS21a | 0.007 | 0.003 |
| *f3* ΔS11a | 0.004 | 0.002 |
| *f1* ΔS11a | 0.004 | 0.001 |
| *f1* ΔS11m | 0.003 | 0.002 |
| *f2* ΔS21a | 0.001 | 0.002 |

Std= standard deviation

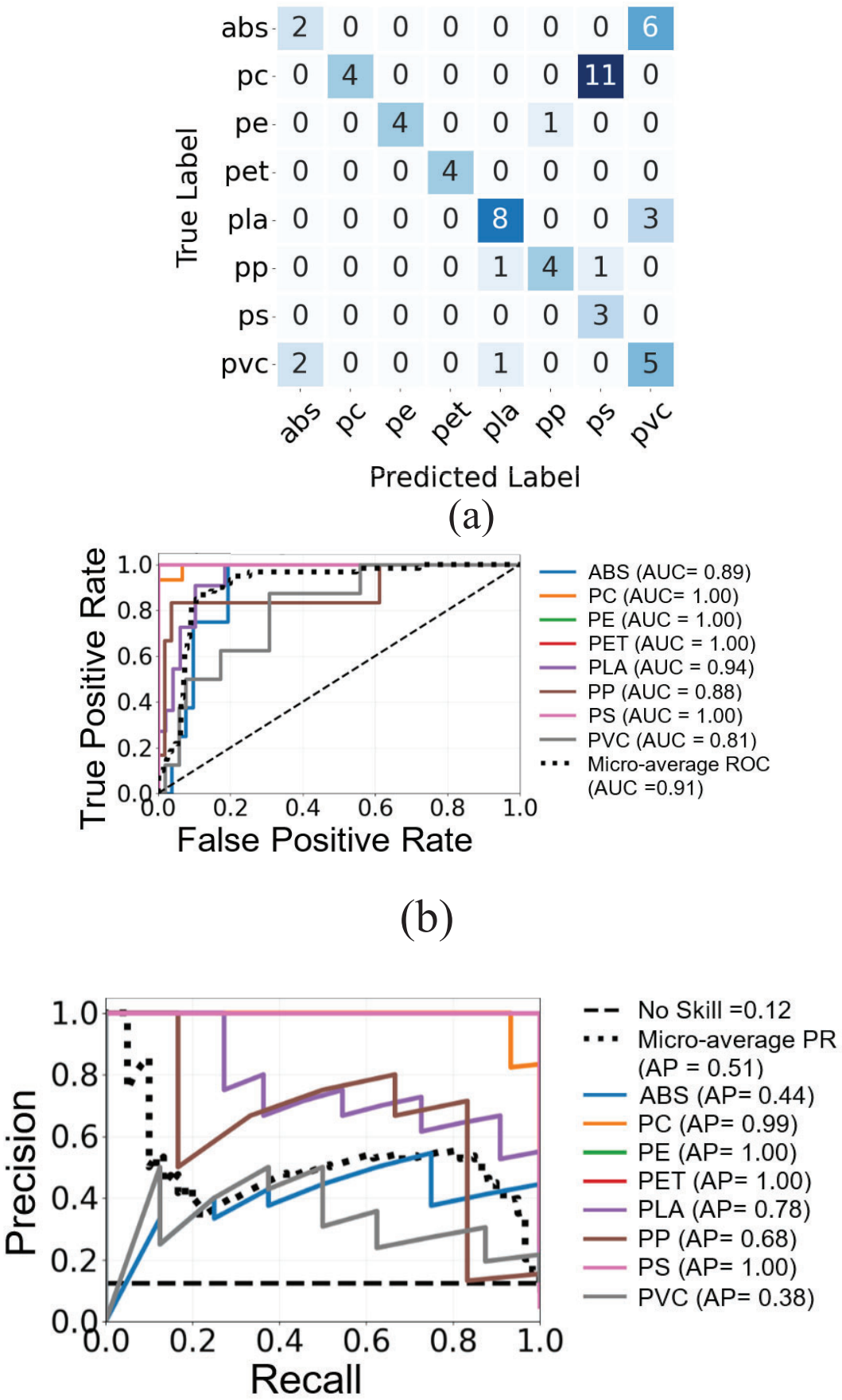

Fig. A6 5 component PCA pipeline classification results majority vote (mAP: .77): (a) confusion matrix (b) ROC curve (c) PR curves

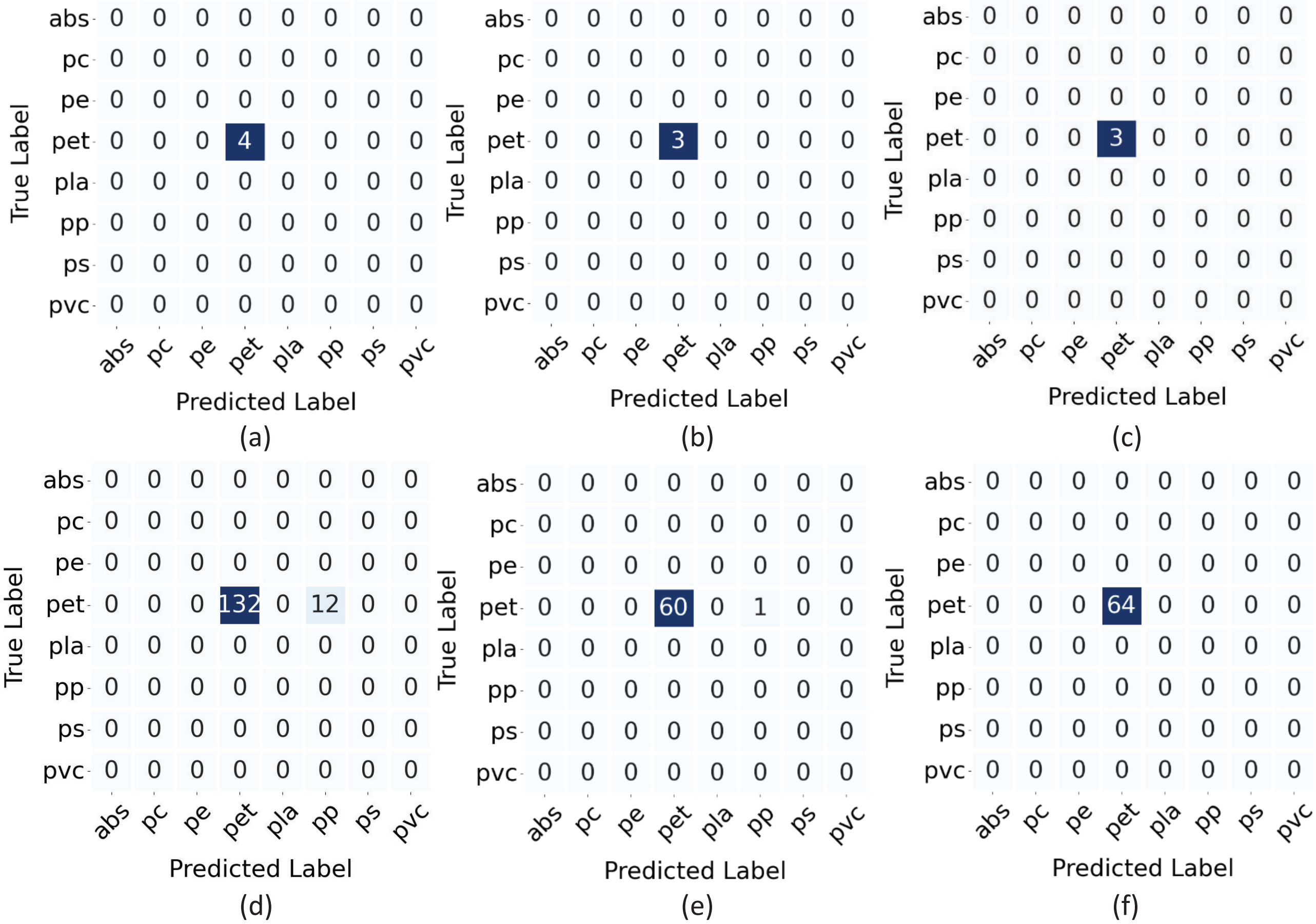

Fig. A7 Majority vote per particle salt test results (a) 0% (b) 3.3% (c) 6.6% sea salt. Individual sample test result (d) 0% (e) 3.3% (f) 6.6% sea salt

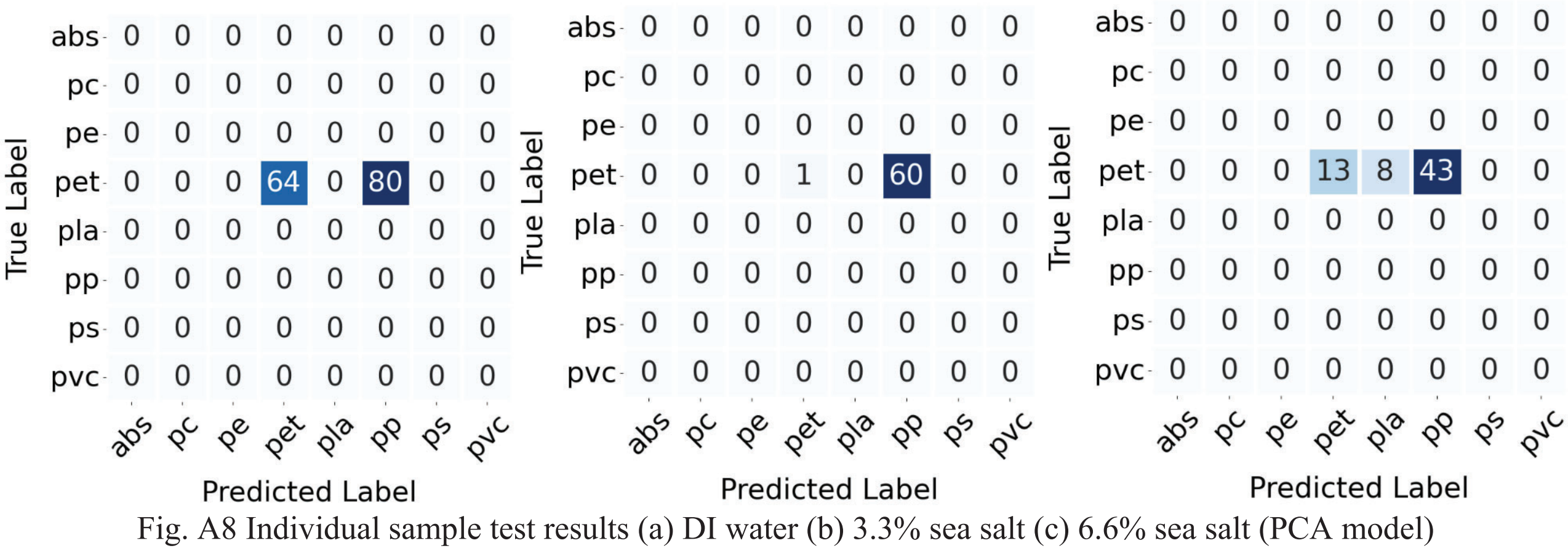

Fig. A8 Individual sample test results (a) DI water (b) 3.3% sea salt (c) 6.6% sea salt (PCA model)